\documentclass[journal]{IEEEtran}
\ifCLASSINFOpdf
   \usepackage[pdftex]{graphicx}
   \graphicspath{{Images/}}
\else
\fi
\usepackage{subcaption}
\begin{document}
%
\title{Light Field Super Resolution Through\\ Controlled Micro-Shifts of Light Field Sensor}
%
%
%

\author{M.~Umair~Mukati
        and~Bahadir~K.~Gunturk%
\thanks{M. Umair Mukati and Bahadir K. Gunturk are with Istanbul Medipol University, Istanbul, 34810 Turkey. E-mail: {umairmukati@gmail.com}, {bkgunturk@medipol.edu.tr}. This work is supported by TUBITAK Grant 114E095.}
}

%
%

\markboth{}%
{Mukati \MakeLowercase{\textit{et al.}}: Light Field Super Resolution Through Controlled Micro-Shifts of Light Field Sensor}
%



\maketitle

\begin{abstract}
Light field cameras enable new capabilities, such as post-capture refocusing and aperture control, through capturing directional and spatial distribution of light rays in space. Micro-lens array based light field camera design is often preferred due to its light transmission efficiency, cost-effectiveness and compactness. One drawback of the micro-lens array based light field cameras is low spatial resolution due to the fact that a single sensor is shared to capture both spatial and angular information. To address the low spatial resolution issue, we present a light field imaging approach, where multiple light fields are captured and fused to improve the spatial resolution. For each capture, the light field sensor is shifted by a pre-determined fraction of a micro-lens size using an {\it xy} translation stage for optimal performance. 
\end{abstract}


%
\IEEEpeerreviewmaketitle

\section{Introduction}
%
%
%
%
\IEEEPARstart{I}{n} light field imaging, the spatial and directional distribution of light rays is recorded in contrast to a regular imaging system, where the angular information of light rays is lost. Light field imaging systems present new capabilities, such as post-capture refocusing, post-capture aperture size and shape control, post-capture viewpoint change, depth estimation, 3D modeling and rendering. The idea of light field imaging is introduced by Lippmann \cite{lippmann1908epreuves}, who proposed to use an array of lenslets in front of a film to capture light field, and used the term ``integral photography" for the idea. The term ``light field" was coined by Gershun \cite{gershun1939light}, who studied the distribution of light in space. Adelson \emph{et al.} \cite{adelson1991plenoptic} conceptualized light field as the entire visual information, including spectral and temporal information in addition to 3D position and angular distribution of light rays in space. In 1996, Levoy \emph{et al.} \cite{levoy1996light} and Gortler \emph{et al.} \cite{gortler1996lumigraph} formulated light field as a four-dimensional function at a time instant, assuming lossless medium.  Since then, light field imaging has become popular, resulting in new applications and theoretical developments. Among different applications of light field imaging are 3D optical inspection, 3D particle flow analysis, object recognition and video post-production. 

There are different ways of capturing light field, including camera arrays \cite{levoy1996light,yang2002real}, micro-lens arrays \cite{ng2006digital,lumsdaine2009focused}, coded masks \cite{veeraraghavan2007dappled}, lens arrays \cite{georgiev2006spatio}, gantry-mounted cameras \cite{unger2003capturing}, and specially designed optics \cite{manakov2013reconfigurable}.  Among these different approaches, micro-lens array (MLA) based light field cameras provide a cost-effective and compact approach to capture light field. There are commercial light field cameras based on the MLA approach \cite{lytro,raytrix}. The main problem with MLA based light field cameras is low spatial resolution due to the fact that a single image sensor is shared to capture both spatial and directional information. For instance, the first generation Lytro camera incorporates a sensor of 11 megapixels but has an effective spatial resolution of about 0.1 megapixels, as each micro-lens produces a single pixel of a light field perspective image \cite{dansereau2013decoding}. In addition to directly affecting depth estimation accuracy \cite{Hahne2016,Gul2018}, low spatial resolution limits the image quality and therefore, widespread application of the light field cameras. 

One possible approach to address the low spatial resolution issue of MLA based light field cameras is to apply a super-resolution restoration method to light field perspective images. Multi-frame super-resolution methods require the input images to provide pixel samples at proper spatial sampling locations. For example, to double the resolution of an image in both horizontal and vertical directions, there should be exactly four input images with half pixel shifts in horizontal, vertical, and diagonal directions. Typically, however, it is necessary to have more than the minimum possible number of input images since the ideal shifts are not guaranteed. The number of input images, along with the need to estimate the shift amounts, add to the computational cost of multi-frame super-resolution methods.  

By controlling the sensor shifts, and designing them for the proper sampling locations, it is possible to achieve the needed resolution enhancement performance with the minimum number of input images. The idea of shifting the sensor in sub-pixel units and compositing a higher resolution image from the input images is known as ``micro-scanning" \cite{lenz1994new}. 

In this paper, we demonstrate the use of the micro-scanning idea for light field spatial resolution enhancement. The shift amounts are designed for a specific light field sensor (the first generation Lytro sensor) considering the arrangement and dimensions of the micro-lenses. We show both qualitatively and quantitatively that, with the application of micro-scanning, the spatial resolution of Lytro light field sensor can be improved significantly. 

In Section \ref{sec:RelatedWork}, we survey the literature related to spatial resolution enhancement of light field cameras and discuss the micro-scanning technique. In Section \ref{sec:Idea}, we give an overview of our approach to improve the spatial resolution of light field images. The prototype imaging system is explained in Section \ref{sec:Prototype}. The details of our resolution enhancement algorithm is given in Section \ref{sec:Reconstruct}. We provide experimental results in Section \ref{sec:Results}. And finally, we conclude our paper in Section \ref{sec:Conclusion}.

\section{Related work}
\label{sec:RelatedWork}
\subsection{Light field spatial resolution enhancement}

There are a number of software based methods proposed to improve the spatial resolution light field perspective images, including frequency domain interpolation \cite{levin2010linear}, Bayesian super-resolution restoration with texture priors \cite{bishop2012light}, Gaussian mixture modeling of high-resolution patches \cite{mitra2012light}, dictionary learning \cite{cho2013modeling}, variational estimation \cite{wanner2014variational}, and deep convolutional neural network \cite{yoon2015learning}. These methods can be applied to both classical light field cameras \cite{ng2005light}, where the objective lens forms the image on the MLA, as well as the ``focused" light field cameras \cite{lumsdaine2009focused}, where micro-lenses and objective lens together form the image on the sensor. (In a standard light field camera, the pixels behind a micro-lens captures the angular information, while the number of micro-lenses forms the spatial resolution of the camera. In a focused light field camera, each micro-lens forms a perspective image on the sensor, that is, the number of pixels behind a micro-lens forms the spatial resolution, while the number of micro-lenses forms the angular resolution.) There are also methods specifically designed to improve the spatial resolution of focused light field cameras \cite{georgiev2009superresolution}.  

Hybrid systems that include a light field camera and a conventional camera have also been proposed to improve the spatial resolution light fields. In \cite{boominathan2014improving}, high-resolution patches from the conventional camera are used to improve the spatial resolution light field perspective images. A dictionary learning method is presented in \cite{wu2015novel}. In \cite{alam2016hybrid}, optical flow based registration is used for efficient resolution enhancement. When the light field camera and the conventional camera do not have the same optical axis, the cameras have different viewpoints, resulting in occluded regions; and the above-mentioned methods should be able to tackle this issue. In \cite{wang2016high} and \cite{dai2014high}, beamsplitters are used in front of the cameras to have the same optical axis and prevent the occlusion issue. In \cite{mukati2017}, the objective lens is common and the beamsplitter is placed in front of the sensors to avoid any issues (such as different optical distortions) due to mismatching objective lenses. 

\subsection{Micro-scanning}

In multi-frame super-resolution image restoration, multiple input images are fused to increase the spatial resolution. It is required to have proper shifts among the input images (as a result of the camera movement or the movement of objects in the scene) to have the diversity of pixel sampling locations. Instead of capturing large number of images and hoping to have proper movement among these images to cover the pixel sampling space, one can induce the movement on the sensor directly and record the necessary samples with the minimum number of possible image captures. For example, an image sensor can be moved by half a pixel in horizontal, vertical and diagonal directions to produce four different images, which are later composited to form an image with four times the original resolution. This micro-scanning idea \cite{lenz1994new} has been used in digital photography, scanners, microscopy, and video recording applications \cite{ben2004jitter}. One drawback of the micro-scanning technique is that the scene needs to be static during the capture of multiple images in order to avoid any registration process. Using piezo-electric actuators synchronized with fast image sensors, the dynamic scene issue can be alleviated to some degree. Even when there are moving objects in the scene, the technique can be applied through incorporating a registration step into the restoration process, which would take care of the moving regions while ensuring there is the required movement in the static regions. 

There are few papers relevant to bringing the idea of micro-scanning and light field imaging together. In \cite{Jang2002}, two micro-lens arrays (one for pickup and one for display purposes) are used. By changing the position of the lenslets for pickup and display processes within the time constant of the eyes' response time, the viewing resolution is improved.  In \cite{Erdmann2001}, the imaging system includes a micro-lens array and a pinhole array. The relative position of a pinhole with respect to a micro-lens gives a specific viewing position. By sequentially capturing images with different viewing positions, a light field is generated. In other words, micro-scanning is used to generate a light field whose angular resolution is determined by the number of scans; and the spatial resolution is limited by the number of pixels on the image sensor.

There are two papers closely related to our work, aiming to improve the spatial resolution through micro-scanning. In \cite{Lim2009}, a microscopic light field imaging system, consisting of a macro lens, a micro-lens array (attached to an {\it xy} translation stage), and an objective lens, is presented. The micro-lens array is shifted relative to the image sensor to form higher spatial density elemental images. In \cite{Yang2015}, the micro-lens array is placed next to the object to be imaged. The micro-array lens is shifted by a translation stage to capture multiple images with a regular camera. These images are then fused to obtain a higher spatial resolution light field. The proposed work has has hardware and algorithmic differences than these previous micro-scanning methods. In our optical system, the micro-lens array is attached to the image sensor; this eliminates need for a relay lens (as opposed to the previous methods), potentially leading to a more compact form factor. In \cite{Yang2015}, the micro-lens array is placed next to the object, which is limiting the applicability; while our setup can be adjusted for different photography and machine vision applications by simply changing the objective lens. On the software side, we incorporate registration and deconvolution processes that address potential translation inaccuracies and the point spread function of the imaging system.   

\section{Enhancing spatial resolution through micro-shifted light field sensor}
\label{sec:Idea}

The spatial resolution of an MLA based light field camera is determined by the number of micro-lenses in the micro-lens array. For example, in the first-generation Lytro camera, there are about 0.1 million micro-lenses packed in a hexagonal grid in front of a sensor. There is about a 9x9 pixel region behind each micro-lens, resulting in 9x9 perspective images ($i.e.$, angular resolution). The decoding process includes conversion (interpolation) from the hexagonal grid to a square grid to form perspective images \cite{dansereau2013decoding}. In order to increase the spatial resolution, the micro-lens size could be reduced, increasing the sampling density of the micro-lens array. With denser packing of micro-lenses, the main lens aperture should also be reduced to avoid overlaps of the light rays on the sensors; as a result, the angular resolution is reduced.

\begin{figure}[t]
\centering
	\includegraphics[width=3in]{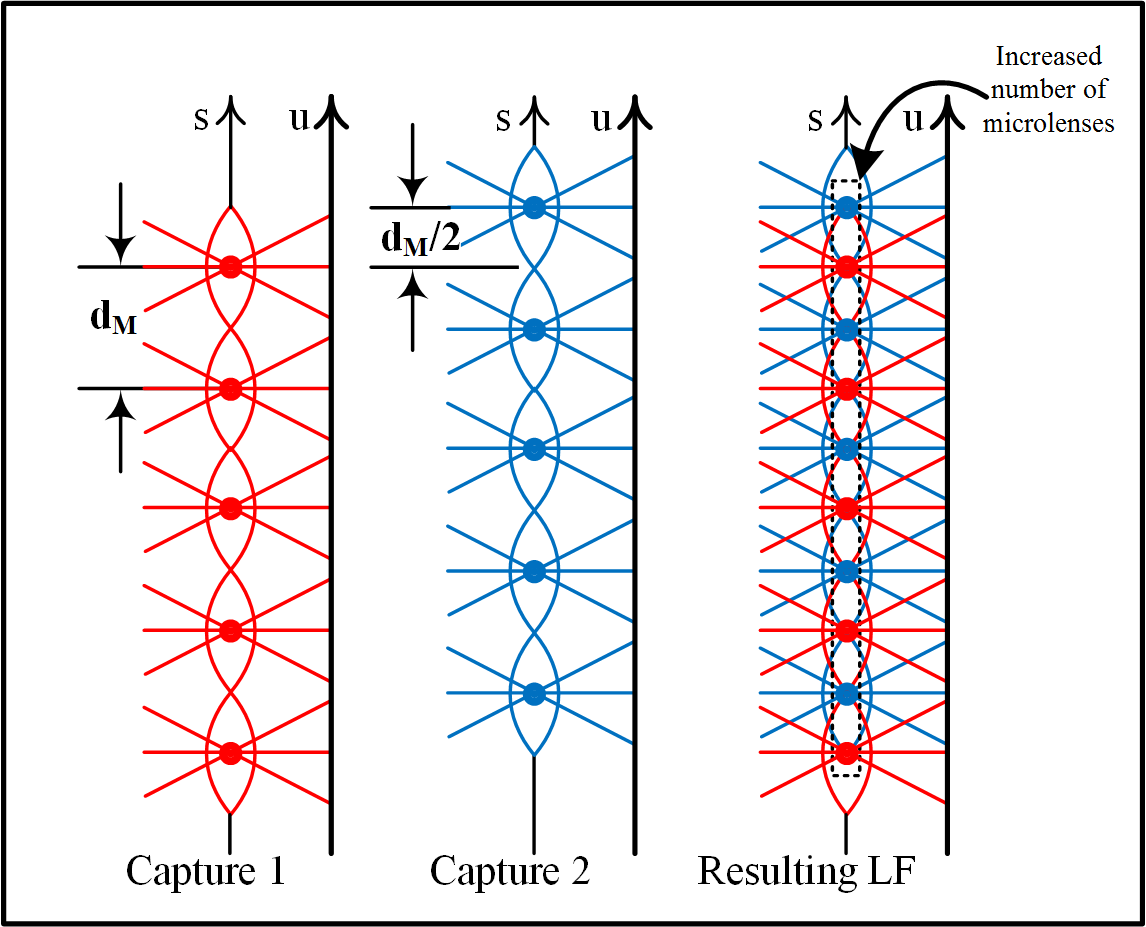}
	\caption{Light field (LF) spatial resolution enhancement by merging multiple light field captures that are shifted with respect to each other. $d_M$ is the distance between the centers of two adjacent micro-lenses; and the shift amount is $d_M / 2$ to double the spatial resolution.}
	\label{fig:Illus}
\end{figure}

In the proposed imaging system, the light rays are recorded at a finer spatial distribution by applying micro-shifts to the light field (LF) sensor. The idea is illustrated for one dimension in Figure \ref{fig:Illus}, where two light fields are captured with a vertical shift of half the size of a micro-lens. The combination of these two light field captures will result in a light field with twice the spatial resolution of the light field sensor in the vertical direction. In other words, the micro-lens sampling density is increased while preserving the angular resolution. 

\section{Camera prototype}
\label{sec:Prototype}

\begin{figure}
\centering
    \includegraphics[width=3in]{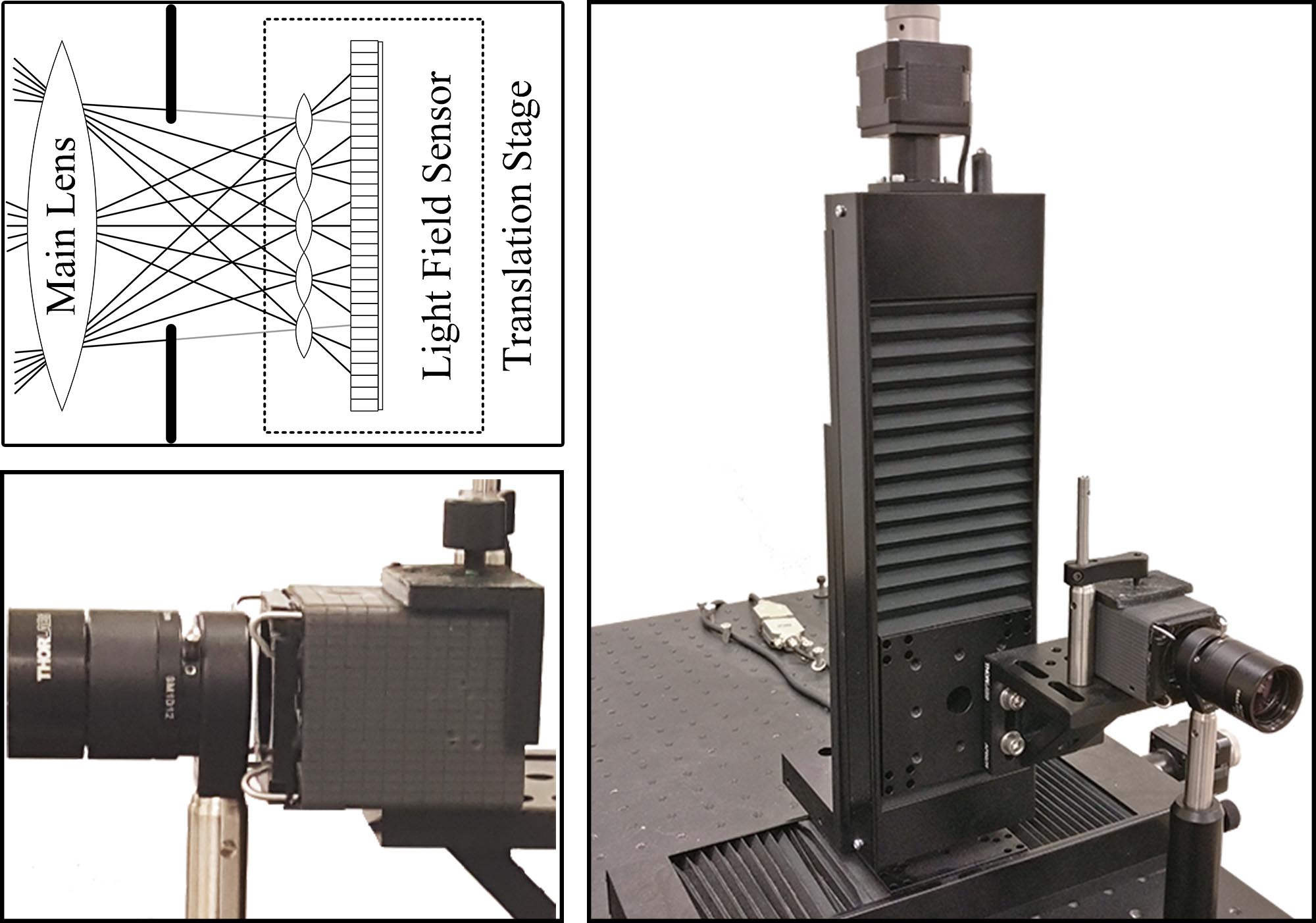}
	\caption{Micro-scanning light field camera prototype. (Top-left) Graphical illustration. The light field sensor is shifted by a translation stage. (Bottom-left) Side view of the optical setup. Note that the lens and the sensor are not joint. (Right) Optical setup showing the translation stages.}
	\label{fig:setup}
\end{figure}

We dismantled a first-generation Lytro camera and removed its objective lens. The sensor part is placed on an {\it xy} translation stage, consisting of two motorized translation stages (specifically, Thorlabs NRT150, which has a minimum achievable incremental movement of 0.1$\mu m$) \cite{thortransstage}. A 60mm objective lens and an optical diaphragm are placed in front of the light field sensor. The sensor and the lens parts are not joined in order to enable independent movement of the sensor. The optical diaphragm is adjusted to match the f-numbers of the objective lens and the micro-lenses. The micro-shifts  are applied to the sensor using the {\it xy} translation stage controlled through the software provided by ThorLabs. The prototype camera is shown in Figure \ref{fig:setup}.

\begin{figure}
\centering
	\includegraphics[width=2in]{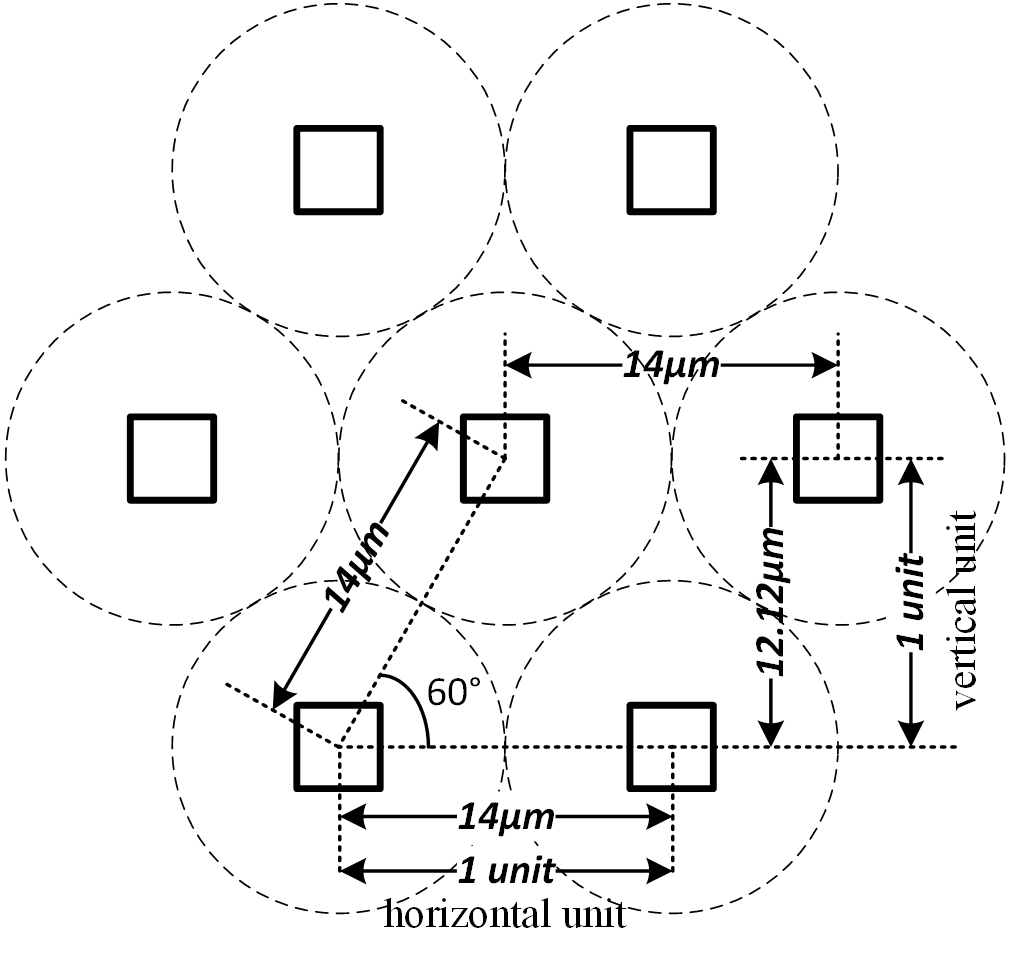}
	\caption{Micro-lens arrangement and dimensions of a first generation Lytro light field sensor.}
	\label{fig:calib}
\end{figure}

The arrangement and dimensions of micro-lenses on the sensor are shown in Figure \ref{fig:calib}. The distance between the centers of two adjacent microlens is 14 $\mu$m; the vertical distance between adjacent two rows of the micro-lens array is 12.12 $\mu$m. The sensor on which the MLA is attached has a size of 3,280 x 3,280 pixels. There are about 0.1 million micro-lenses arranged in a hexagonal pattern.

\begin{figure*}[tbp]
\centering
\begin{subfigure}{1.2in}
  \includegraphics[width=1.2in]{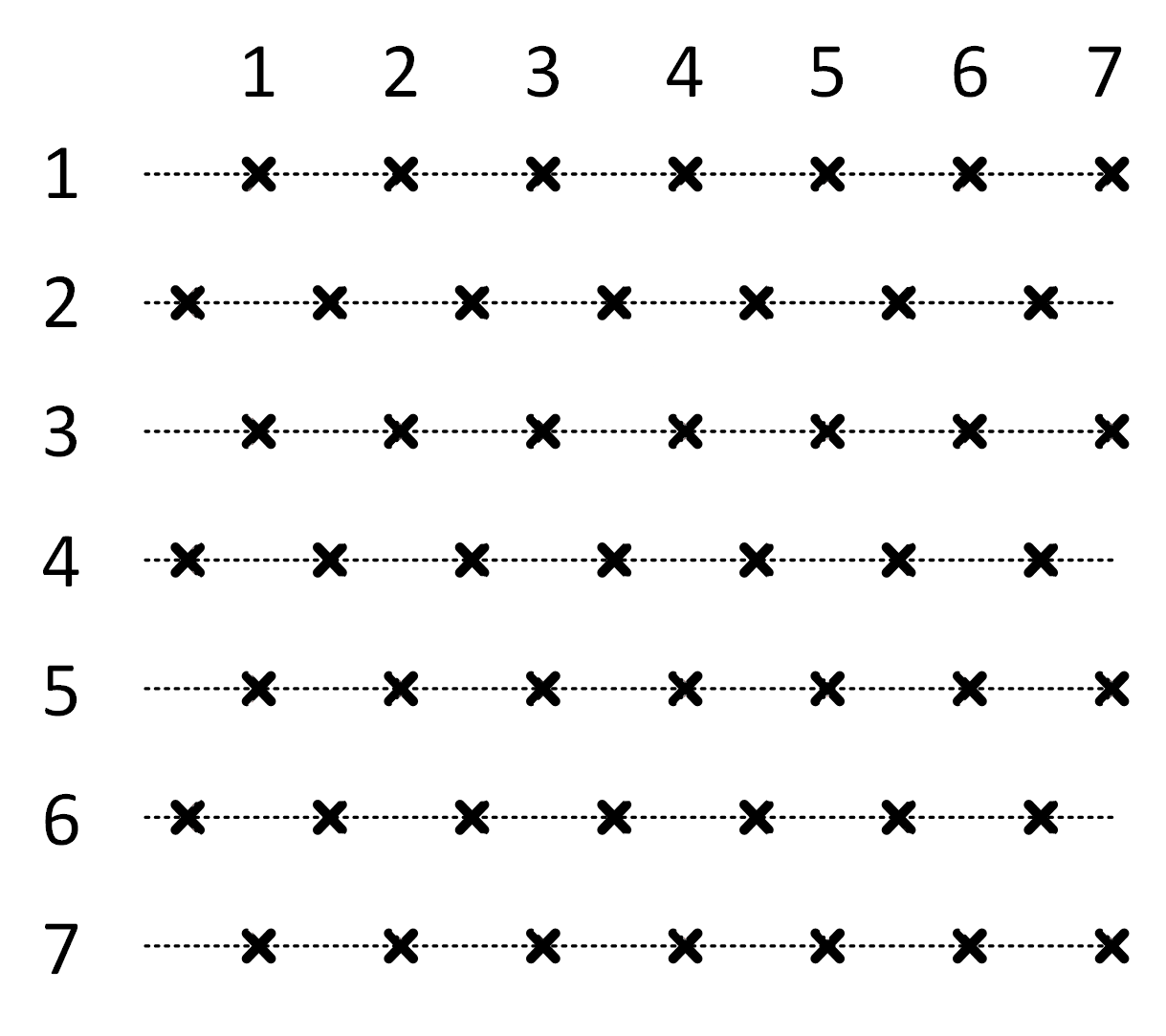}
  \caption{}
  \label{fig:Grid:1}
\end{subfigure}
~
\begin{subfigure}{1.2in}
  \includegraphics[width=1.2in]{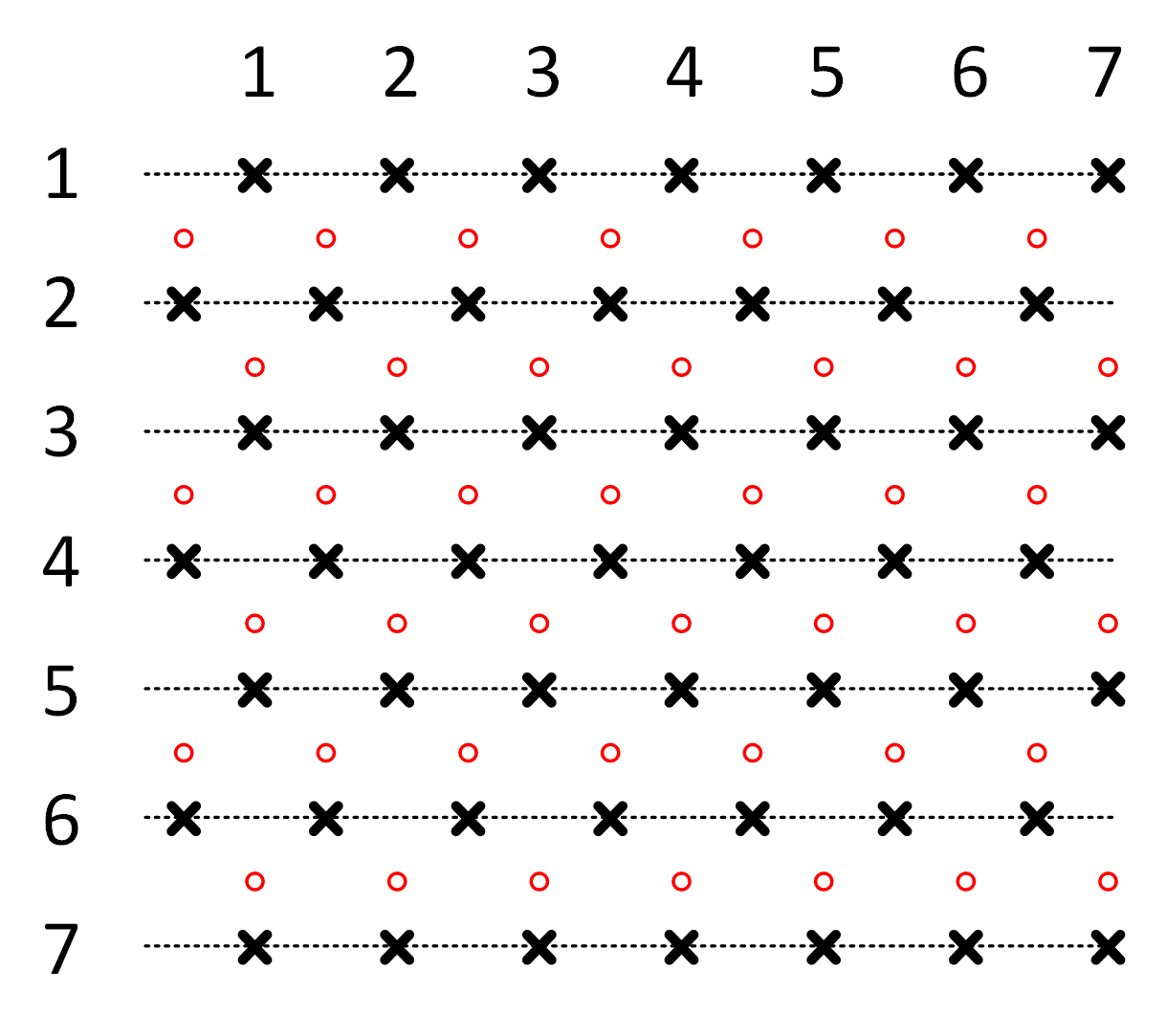}
  \caption{}
  \label{fig:Grid:2}
\end{subfigure}
~
\begin{subfigure}{1.2in}
  \includegraphics[width=1.2in]{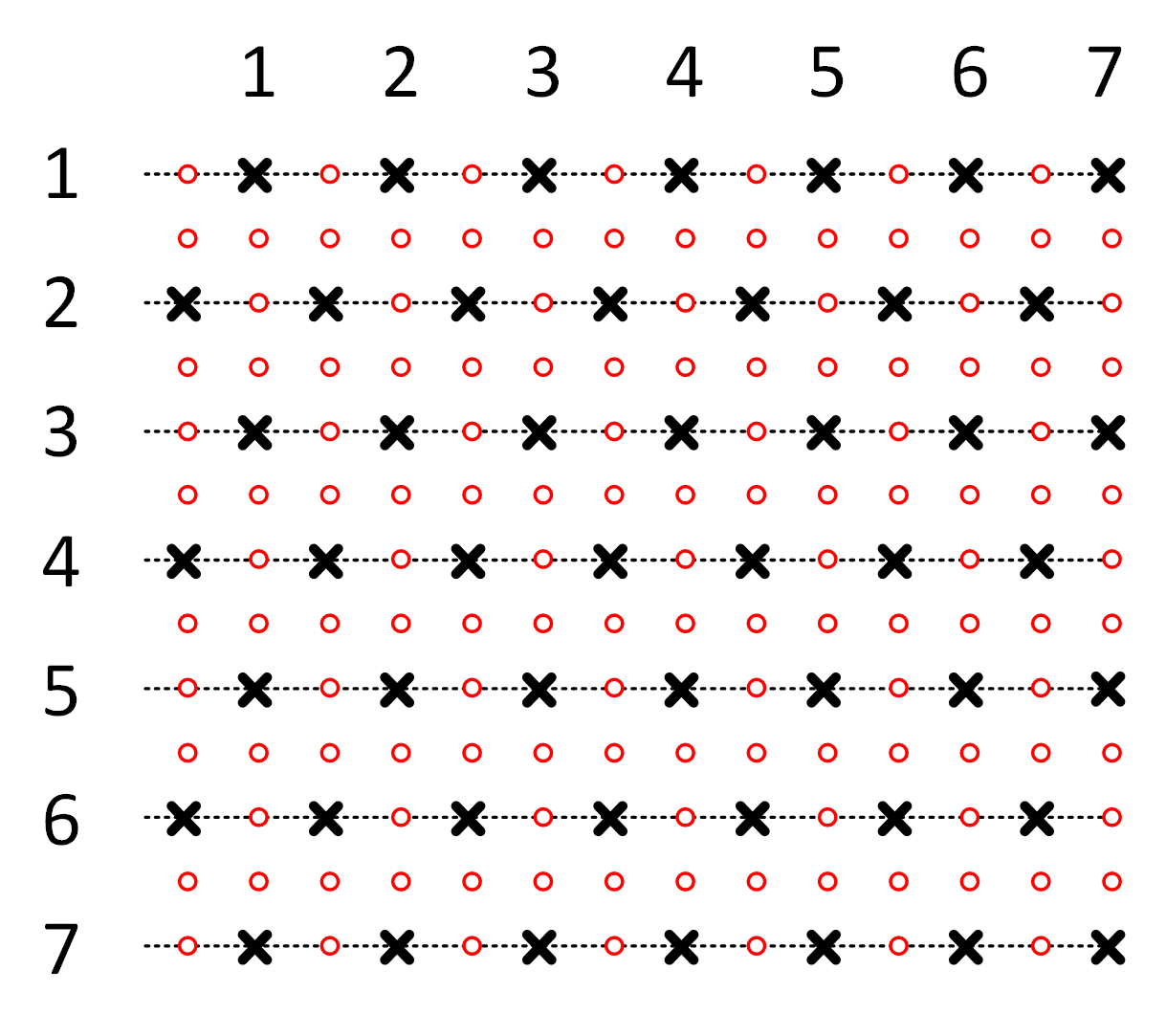}
  \caption{}
  \label{fig:Grid:4}
\end{subfigure}
~
\begin{subfigure}{1.2in}
  \includegraphics[width=1.2in]{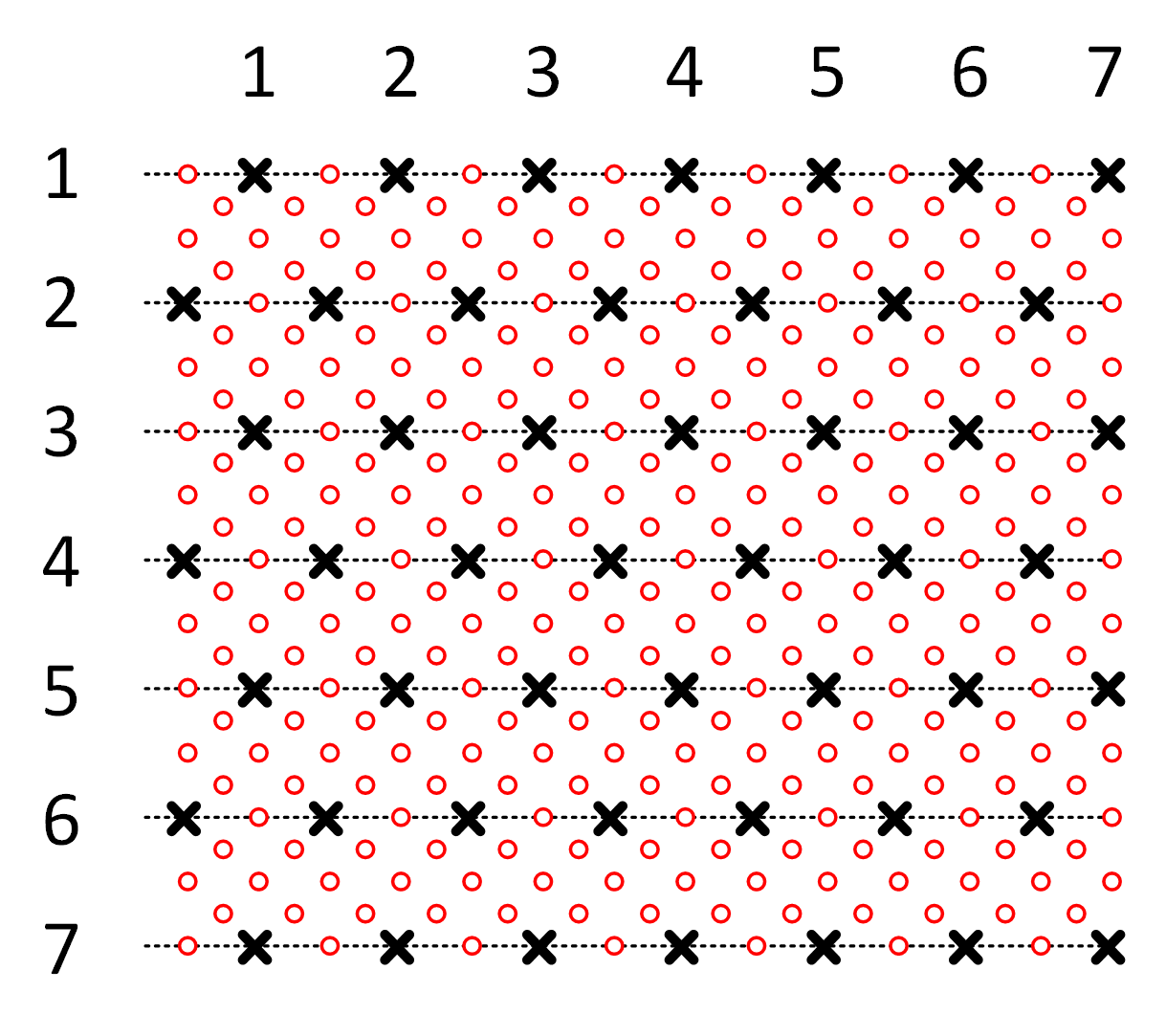}
  \caption{}
  \label{fig:Grid:8}
\end{subfigure}
~
\begin{subfigure}{1.2in}
  \includegraphics[width=1.2in]{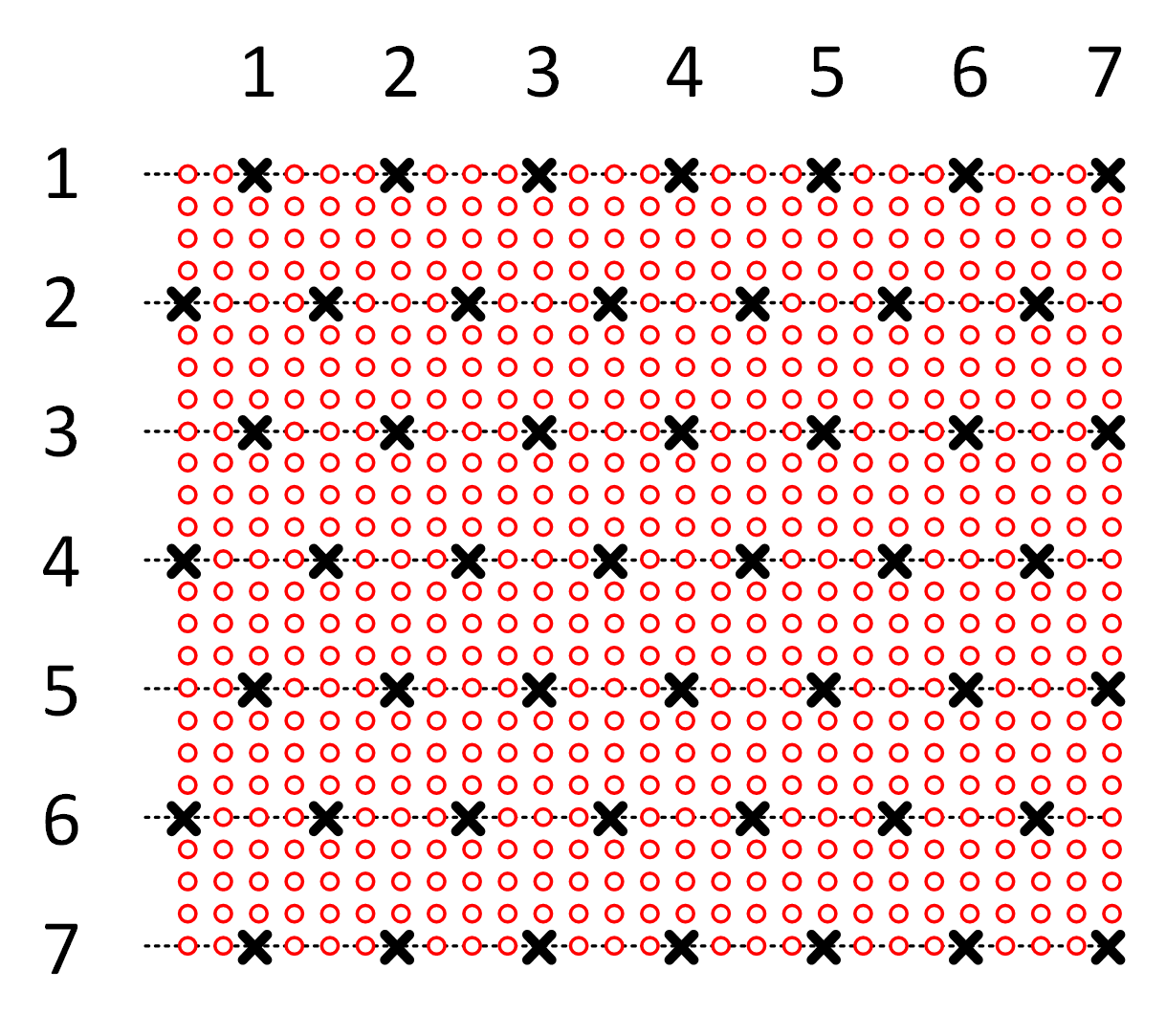}
  \caption{}
  \label{fig:Grid:16}
\end{subfigure}
~  
\caption{Micro-lens locations for various number of input light fields. (a) 1 LF. (b) 2 LFs. (c) 4 LFs. (d) 8 LFs. (e) 16 LFs. }
\label{fig:Grid}
\end{figure*}

We designed the micro-shifts to increase the spatial resolution four times in horizontal direction and four times in vertical direction. We capture 16 light fields; the shift amounts for each capture are listed in Table \ref{tab:1}. The shifts amounts are multiples of the micro-lens spacings divided by four, the resolution enhancement factor along a direction;  that is, the shift amounts are multiples of 14/4=3.5$\mu$m and 12.12/4=3.03 $\mu$m, in horizontal and vertical directions, respectively. For later comparison and analysis of resolution enhancement, additional subsets of light field captures are generated by picking one, two, four and eight light fields from the original 16 captures. All the light field capture sets are shown in Figure \ref{fig:Grid}. The micro-lens grid for a single light field is shown in Figure \ref{fig:Grid:1}, while the grid for 16 light fields is shown in Figure \ref{fig:Grid:16}.

\begin{table}
\centering
\caption{Translation amounts (ideal) in horizontal ($T_u$) and vertical ($T_v$) directions for 16 light field captures. (Refer to Figure \ref{fig:Grid:16} for a graphical illustration.)}
{
\begin{tabular}{|c|rr|c|rr|}
\hline 
Capture & $T_u (\mu m)$ & $T_v (\mu m)$ & Capture & $T_u (\mu m)$ & $T_v (\mu m)$ \\ 
\hline 
1 & 0.00 & 0.00 & 9 & 3.50 & -21.21 \\ 
2 & 0.00 & -3.03 & 10 & 3.50 & -18.18 \\ 
3 & 0.00 & -6.06 & 11 & 3.50 & -15.15 \\ 
4 & 0.00 & -9.09 & 12 & 3.50 & -12.12 \\ 
5 & 0.00 & -12.12 & 13 & 3.50 & -9.09 \\ 
6 & 0.00 & -15.15 & 14 & 3.50 & -6.06 \\ 
7 & 0.00 & -18.18 & 15 & 3.50 & -3.03 \\ 
8 & 0.00 & -21.21 & 16 & 3.50 & 0.00 \\ 
\hline 
\end{tabular} 
}
\label{tab:1}
\end{table}

\section{Reconstructing high resolution light field}
\label{sec:Reconstruct}

For the proper decoding of a light field, the center position for each micro-lens region on the image sensor should be determined. A pixel, behind a micro-lens, records a specific light ray (perspective) according to its position relative to the micro-lens center. Since we dismantled the original Lytro camera, we cannot rely on the optical centers provided by the manufacturer to decode the light field. We followed the procedure described in \cite{dansereau2013decoding} to determine the center positions for each micro-lens region, which involves taking the picture of a white scene and determining the brightest point behind each micro-lens region.

Through micro-scanning, a finer resolution sampling of each light field perspective image is going to be achieved. The sampling locations do not have to be on a regularly spaced grid due to possible optical distortions and inaccurate micro-shifts of the translation stage. To estimate the exact translation between two light fields, we use phase correlation between corresponding perspective images of the two light fields (for example, between the center perspective image of the first light field and the center perspective image of the second light field) to estimate the shift amount. Since the shift between each corresponding perspectives should be equal, we average the estimated shifts to have a robust estimate of the shift between two light fields. The registration process is repeated between a reference light field and the rest of all light field captures. After determining the shift amounts, the recorded samples are interpolated to a finer resolution regularly spaced grid using the Delaunay triangulation based interpolation method \cite{lertrattanapanich2002high}.

The interpolation process explained so far does not take into account the point spread function (PSF) of the imaging system. The objective lens and the micro-lenses obviously contribute to the optical blur at each pixel. In the image super-resolution literature \cite{Gunturk2010chapter}, two main approaches can be adopted. The first approach is to incorporate blurring and warping into the forward model, and recover the high-resolution image iteratively. The second approach is to register the input images and interpolate to a finer resolution grid, followed by a deconvolution process. In this paper, we take the second approach. After interpolation, we apply the blind deconvolution method given in \cite{xu2010two} to improve the perspective images.  

\section{Experimental results}
\label{sec:Results}

\begin{figure}[h]
\centering
	\begin{subfigure}{1.5in}
		\includegraphics[width=1.5in]{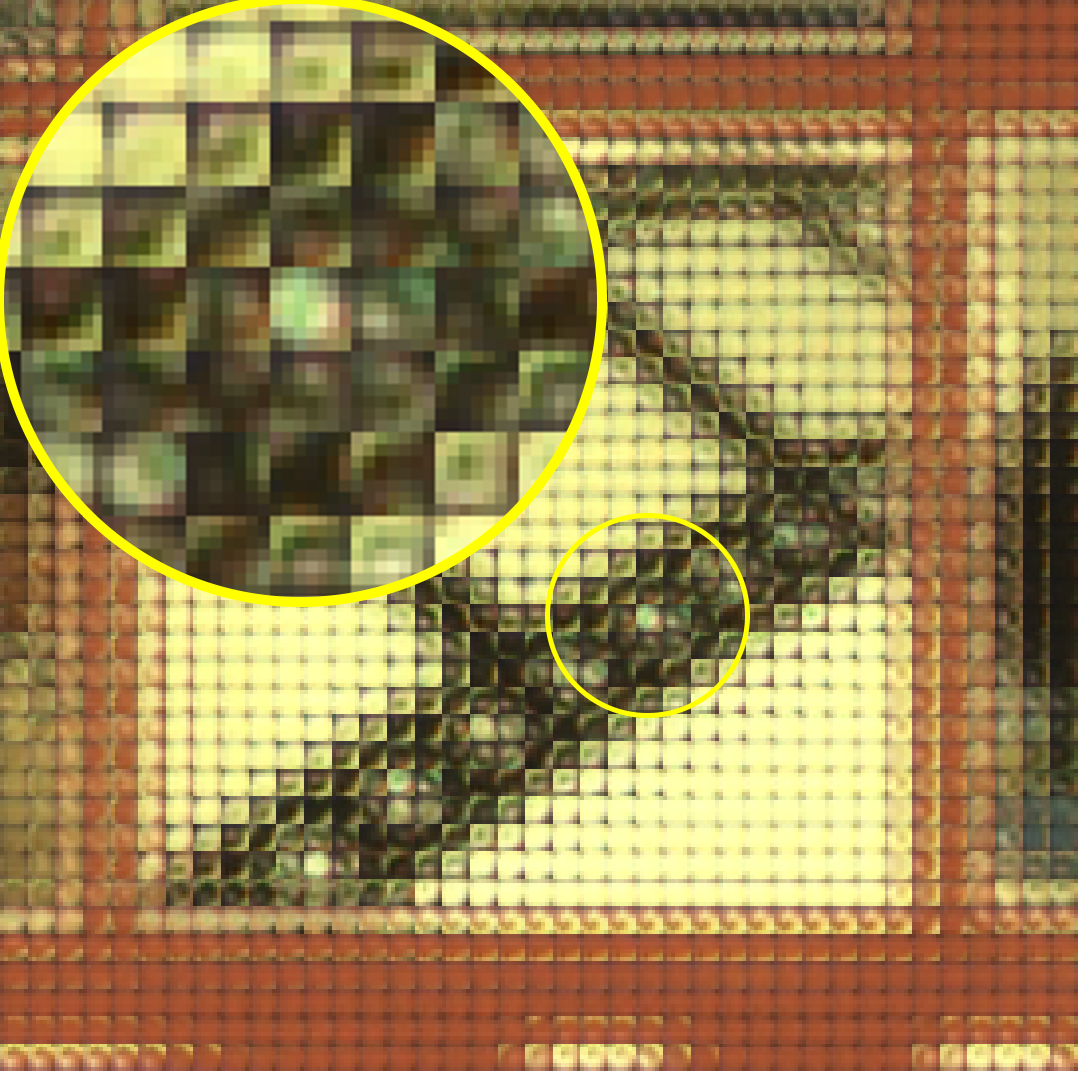}
		\caption{}
		\label{fig:lenslet:small}
	\end{subfigure}
	\begin{subfigure}{1.5in}
		\includegraphics[width=1.5in]{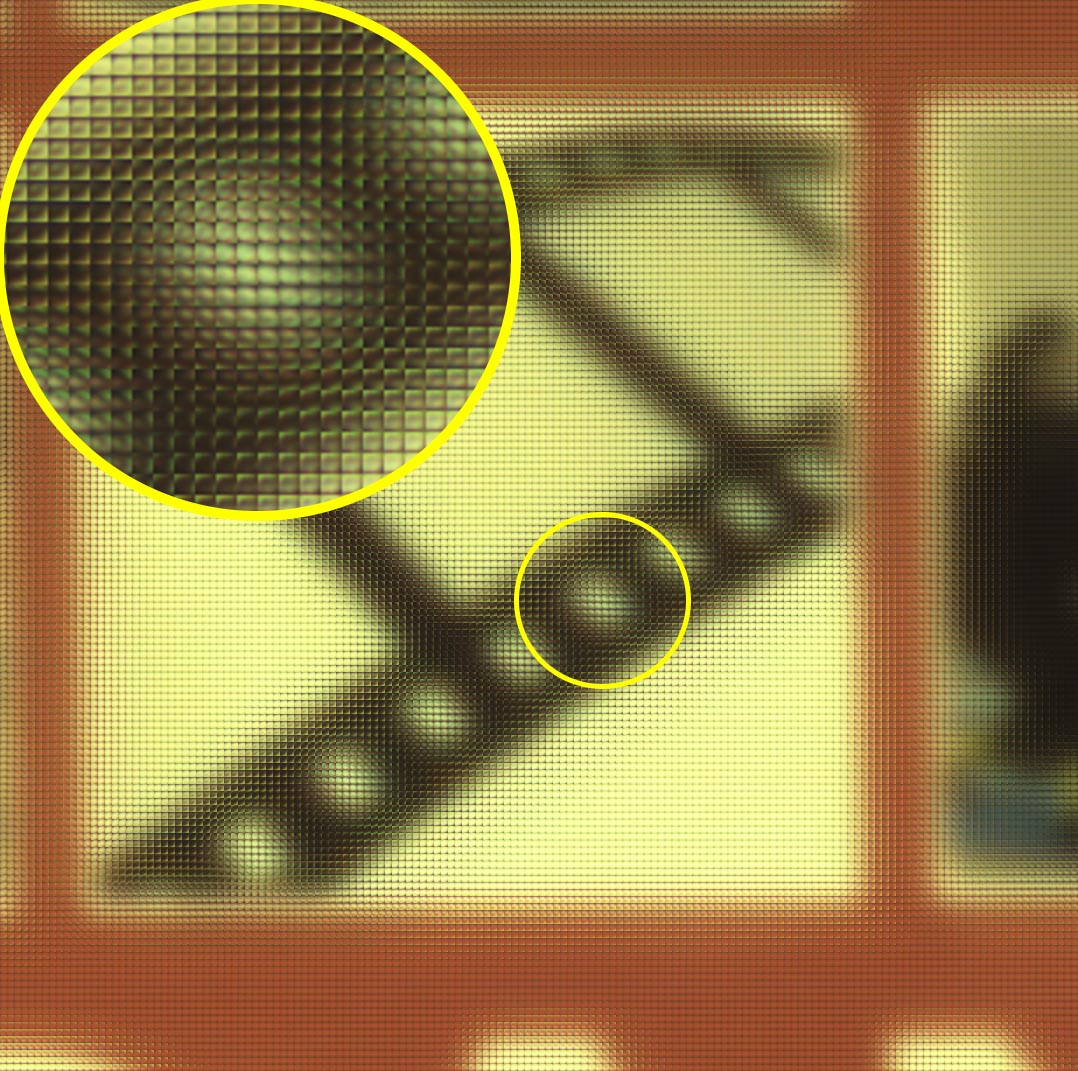}
		\caption{}
		\label{fig:lenslet:large}
	\end{subfigure}
  \caption{Light field spatial resolution enhancement. (a) Low spatial resolution light field. (b) High spatial resolution light field obtained by merging 16 light field captures.}
\label{fig:MicroLens}
\end{figure}

Through the process capturing and fusing multiple (16) light fields, each with spatial resolution of 378x328 pixels and angular resolution of 7x7 perspectives, we obtain a light field of size 1512x1312 pixels, which is 4x4 times than the original spatial resolution. Figure \ref{fig:lenslet:small} shows an input light field, and Figure \ref{fig:lenslet:large} shows the resulting light field at the end of the fusion process.  

\begin{figure*}
\centering
  \includegraphics[width=4.5in]{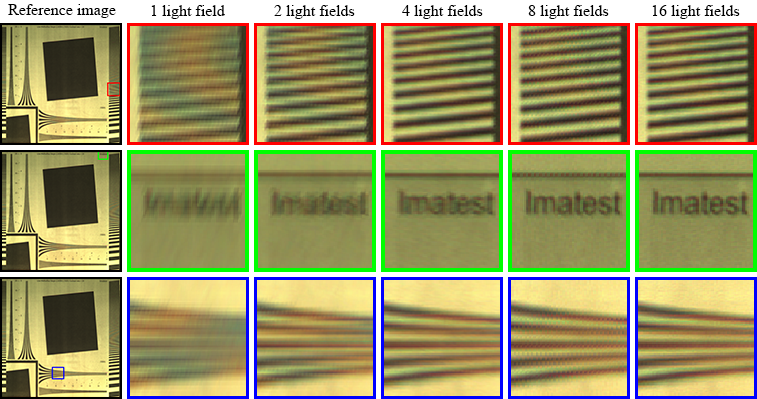}
  \caption{Visual comparison of light field middle perspective images for various number of light field captures used for enhancement.}
\label{fig:Dataset1:numDemo}
\end{figure*}

During the experiment, 16 light fields are captured with the controlled shifts of light field sensor as described in Table \ref{tab:1}. To investigate the effect of the number of light fields, five different subsets of captured light fields are picked as illustrated in Figure \ref{fig:Grid}. For each case, a high resolution light field (with 1512x1312 spatial resolution) is obtained using the interpolation method described in the previous section. In Figure \ref{fig:Dataset1:numDemo}, we provide a visual comparison of the resulting perspective images. As expected, when we use all 16 light fields, we achieve the best visual performance. 

\begin{figure}
\centering
  \includegraphics[width=5.2cm,angle=-90]{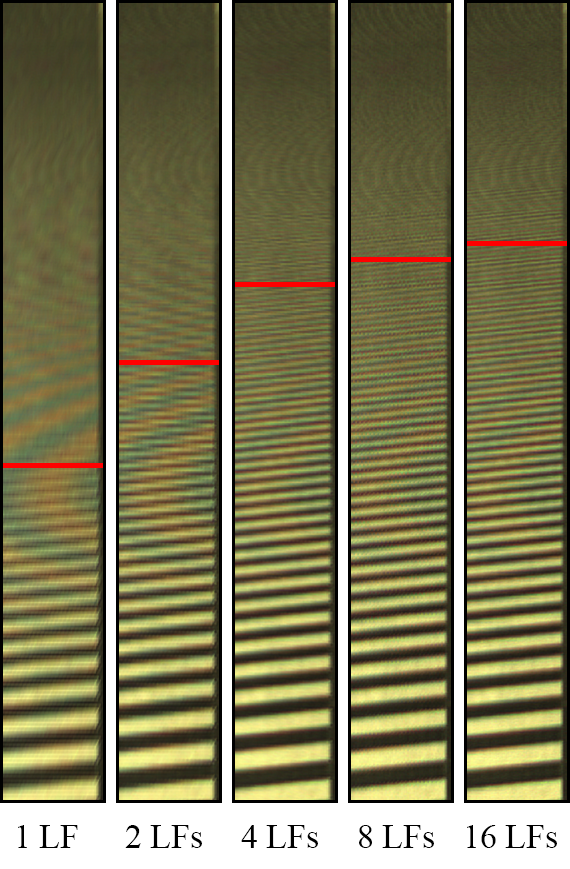}
  \caption{The effect of the number of light field captures used on spatial resolution enhancement.}
\label{fig:Dataset1:testChartCutoff}
\end{figure}

\begin{figure}
\centering
  \includegraphics[width=6.0cm]{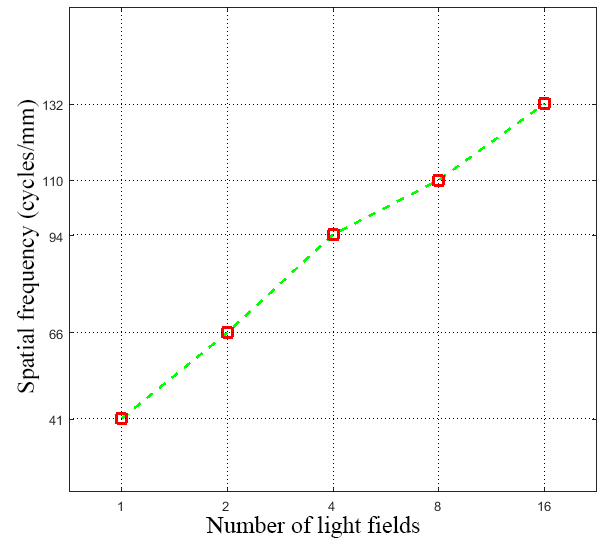}
  \caption{Highest spatial resolution achieved versus the number of light field captures used in light field formation.}
\label{fig:Dataset1:spatRes}
\end{figure}

To quantify the resolution enhancement, we used a region from the test chart, shown in Figure \ref{fig:Dataset1:testChartCutoff}, to determine the highest spatial frequency. The region consists of horizontal bars with increasing spatial frequency (cycles per mm). In each case, we profiled the bars, and marked the location beyond which the bars become indistinguishable through visual inspection; the highest spatial frequencies are plotted in Figure \ref{fig:Dataset1:spatRes}. 

In addition, we obtained the maximum spatial frequencies achieved by \cite{dansereau2013decoding} and \cite{cho2013modeling}  for the same test chart shown in Figure \ref{fig:Dataset1:testChartCutoff}. The spatial frequency for \cite{cho2013modeling} was near 50 cycles/mm and for \cite{dansereau2013decoding} it was around 55 cycles/mm, whereas for resulting light field with 16 captures for interpolation the spatial frequency was around 132 cycles/mm.

\begin{figure*}
\centering
  \rotatebox{90}{~~~~~~~~Before deconvolution}
  \includegraphics[width=5in]{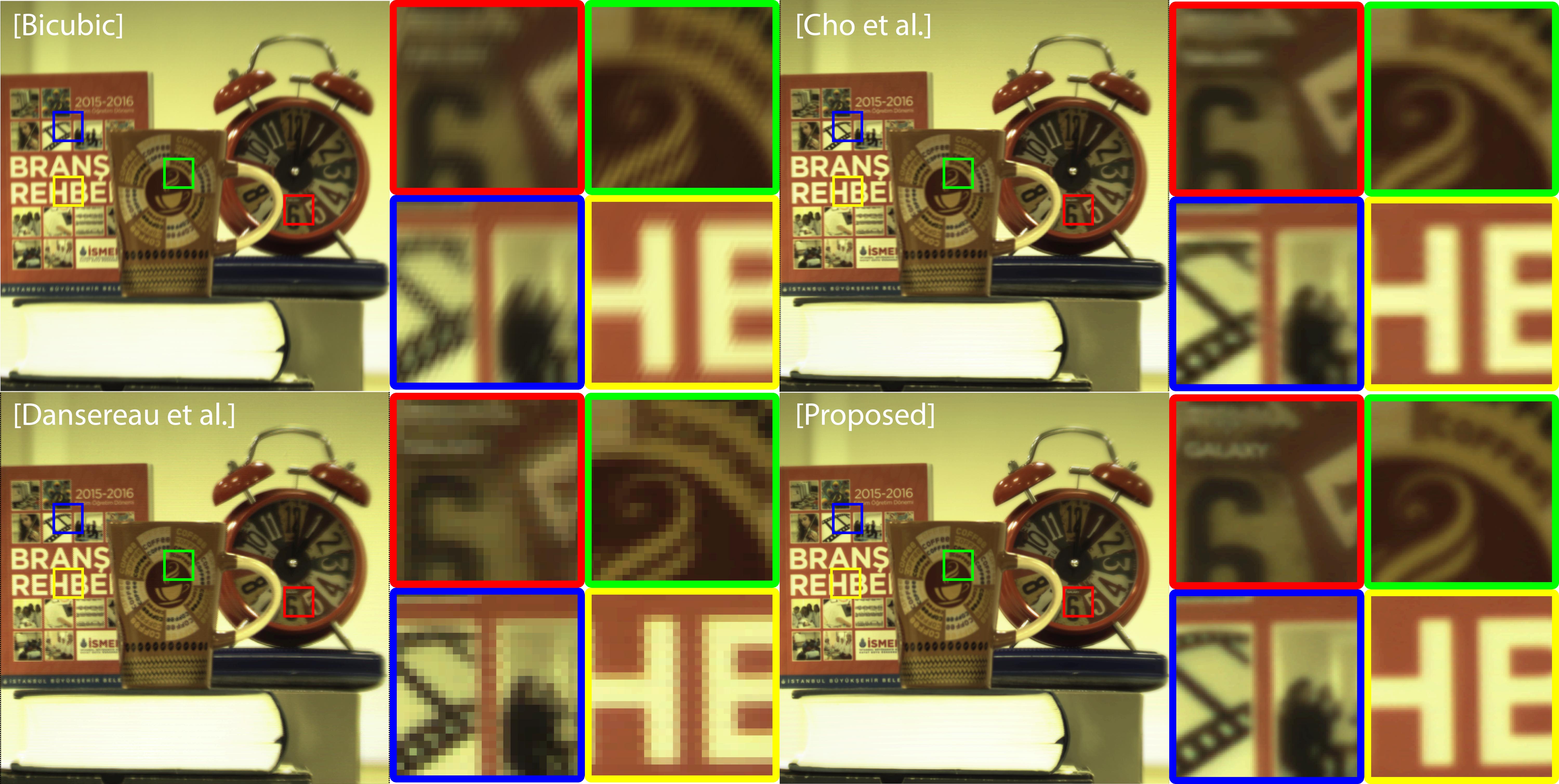}\\
  \vspace{0.1cm}
  \rotatebox{90}{~~~~~~~~After deconvolution}
  \includegraphics[width=5in]{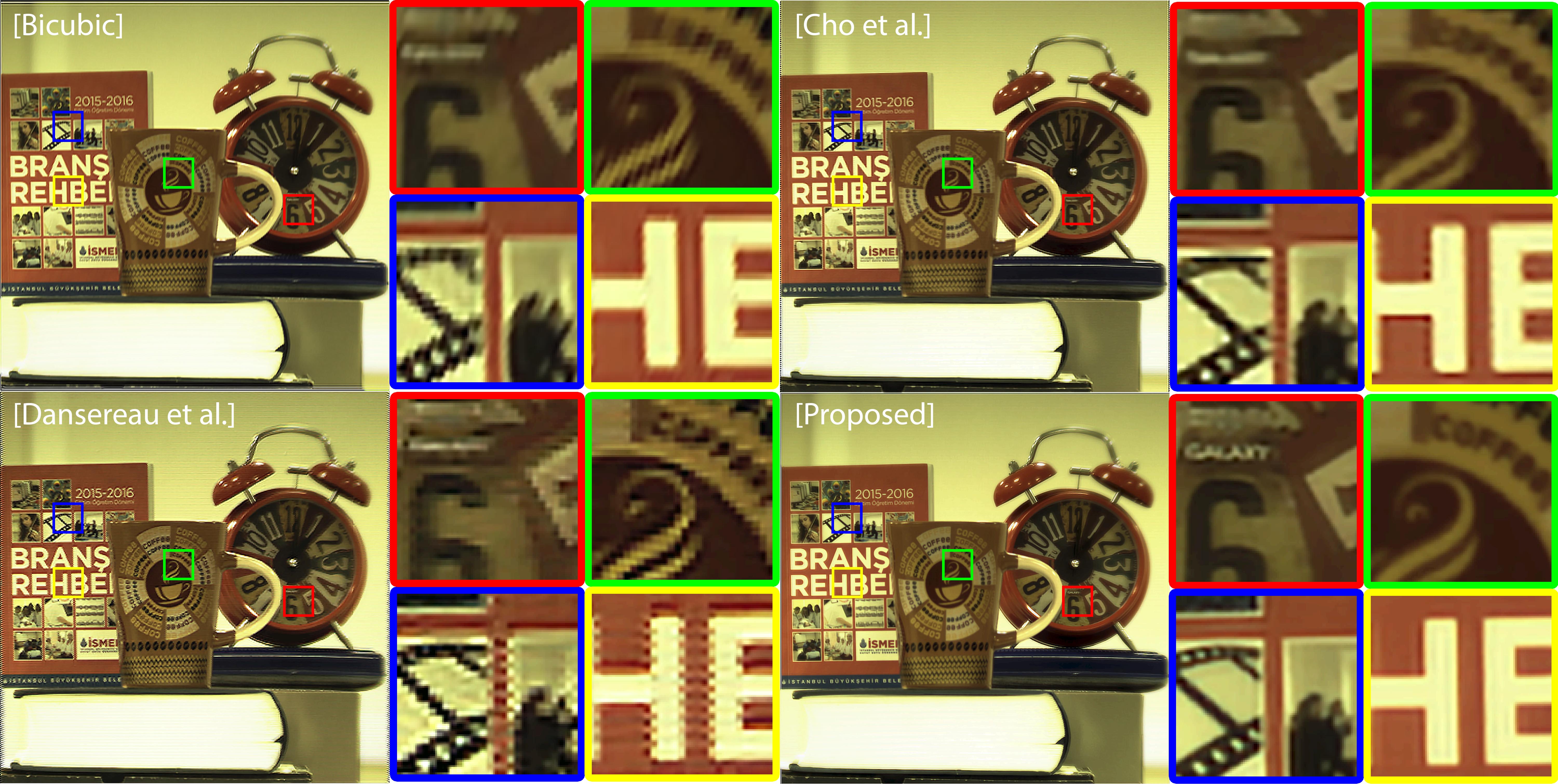}
  \caption{Comparison of light field middle perspective images generated using different techniques, including bicubic interpolation from a single light field capture, light field obtained by Dansereau \emph{et al.} \cite{dansereau2013decoding}, light field obtained by Cho \emph{et al.} \cite{cho2013modeling}, and light field obtained by proposed method merging 16 input light fields.}
\label{fig:Dataset2}
\end{figure*}

\begin{figure*}
\centering
 \rotatebox{90}{~~~~~~~~Before deconvolution}
  \includegraphics[width=5in]{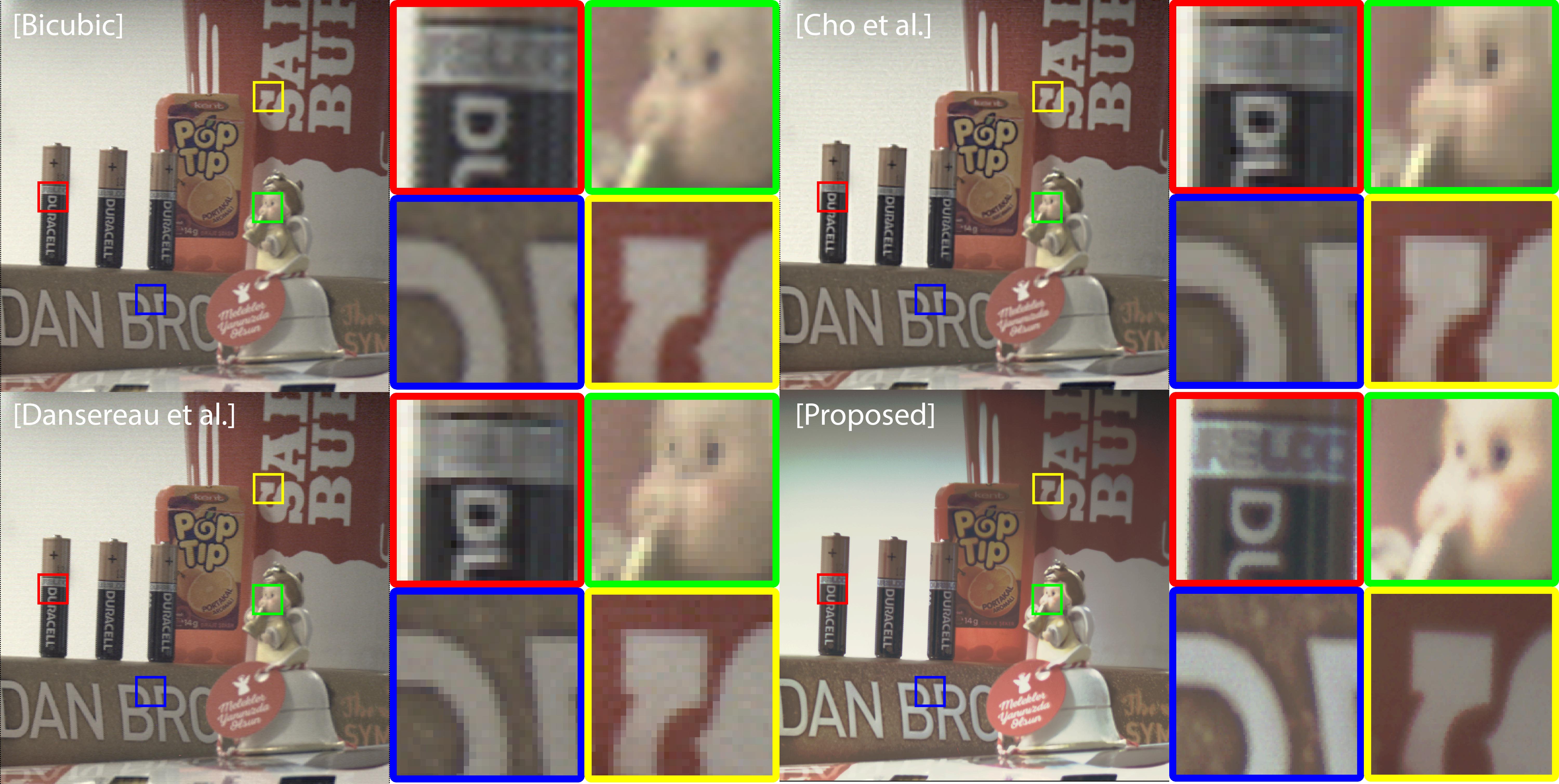}\\
  \vspace{0.1cm}
  \rotatebox{90}{~~~~~~~~After deconvolution}
  \includegraphics[width=5in]{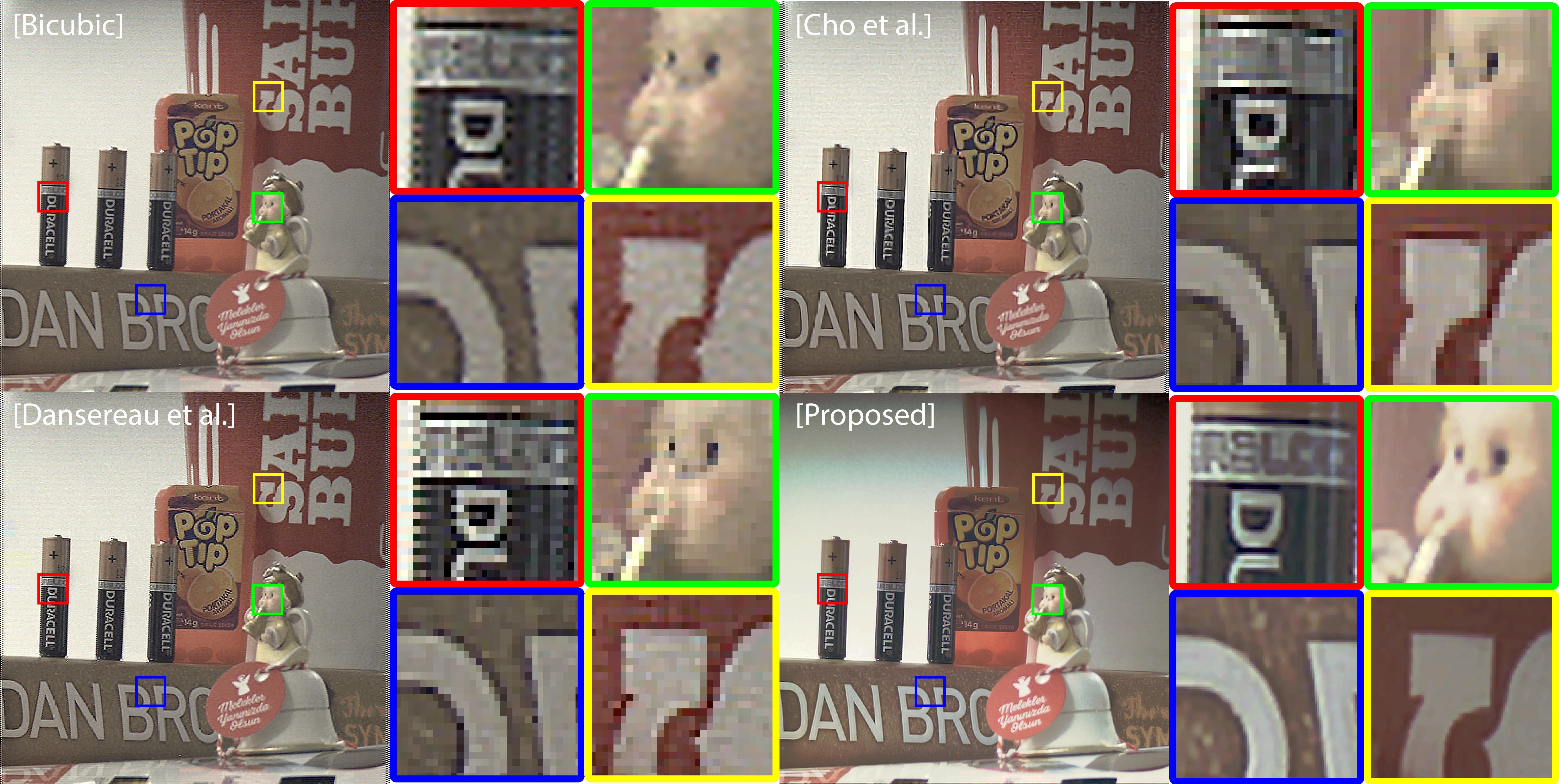}
  \caption{Visual comparison of the middle perspective of the light fields generated using different techniques. Light fields include bicubically interpolated single light field capture, light field obtained by Dansereau \emph{et al.} \cite{dansereau2013decoding}, light field obtained by Cho \emph{et al.} \cite{cho2013modeling}, and light field obtained by proposed method merging 16 input light fields.}
\label{fig:Dataset3}
\end{figure*}

In figures \ref{fig:Dataset2} and \ref{fig:Dataset3}, we present two light field perspective images generated using several techniques, including bicubic interpolation from a single light field capture, the decoding techniques provided in \cite{dansereau2013decoding} and \cite{cho2013modeling}, and the proposed method where 16 light fields are combined after micro-scanning. We present results before and after the application of the blind deconvolution technique \cite{xu2010two}. The resolution enhancement with the proposed micro-scanning approach is obvious in both cases. The proposed method benefits from the deconvolution process, producing sharper results, while the images with the other techniques may deteriorate since the deconvolution process amplifies the artifacts of the decoded image. 

\begin{figure*}
\centering
\rotatebox{90}{~~~Bicubic (before deconv.)}
  \includegraphics[width=4cm]{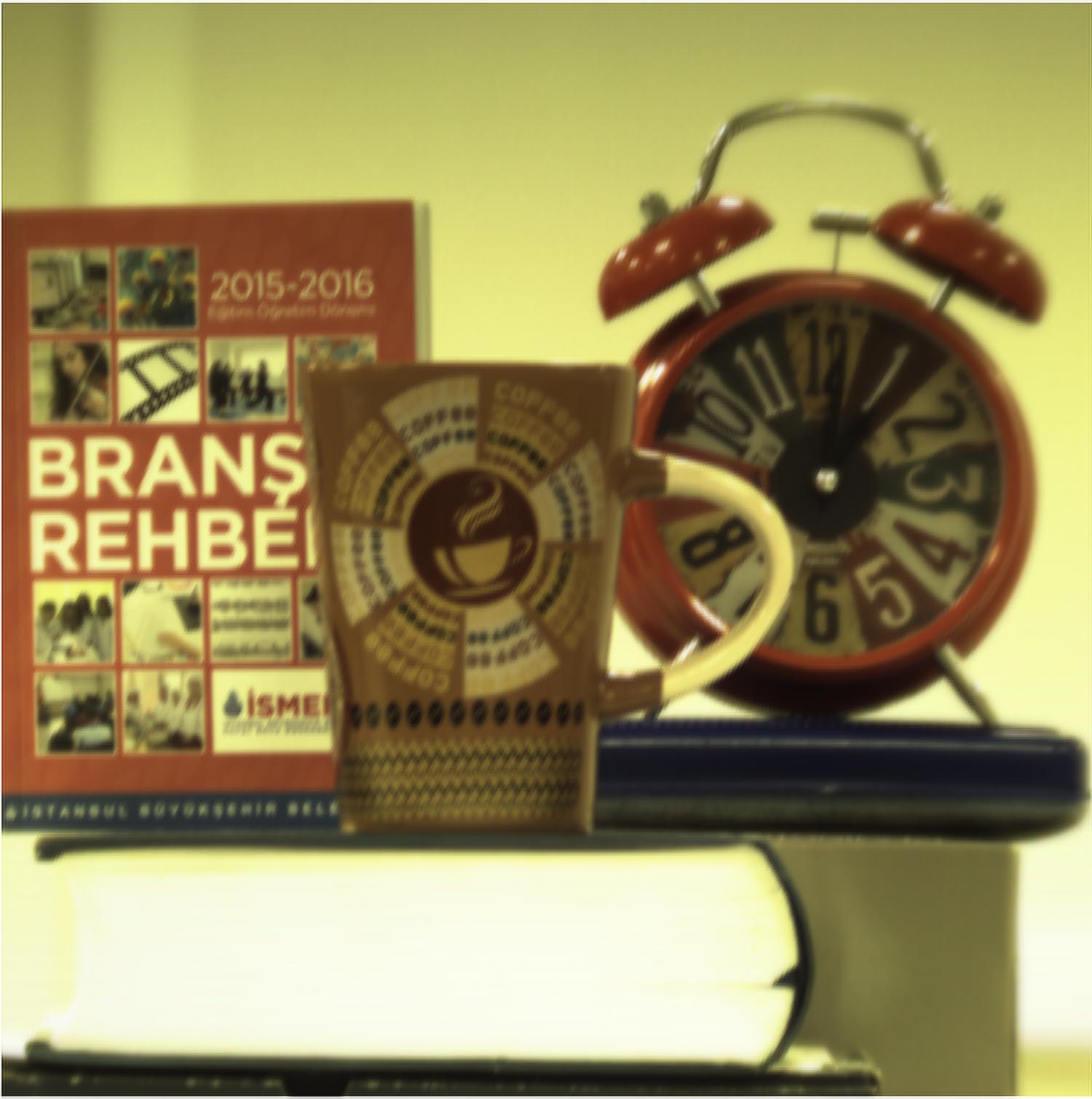}
  \includegraphics[width=2cm,trim={100 850 1100 350},clip=true,width=3.2cm,keepaspectratio]{RF_Mid_Full.jpg}
  \includegraphics[width=2cm,trim={500 800 800 500},clip=true,width=3.2cm,keepaspectratio]{RF_Mid_Full.jpg}  
  \\
  \rotatebox{90}{~~~Bicubic (after deconv.)}
  \includegraphics[width=4cm]{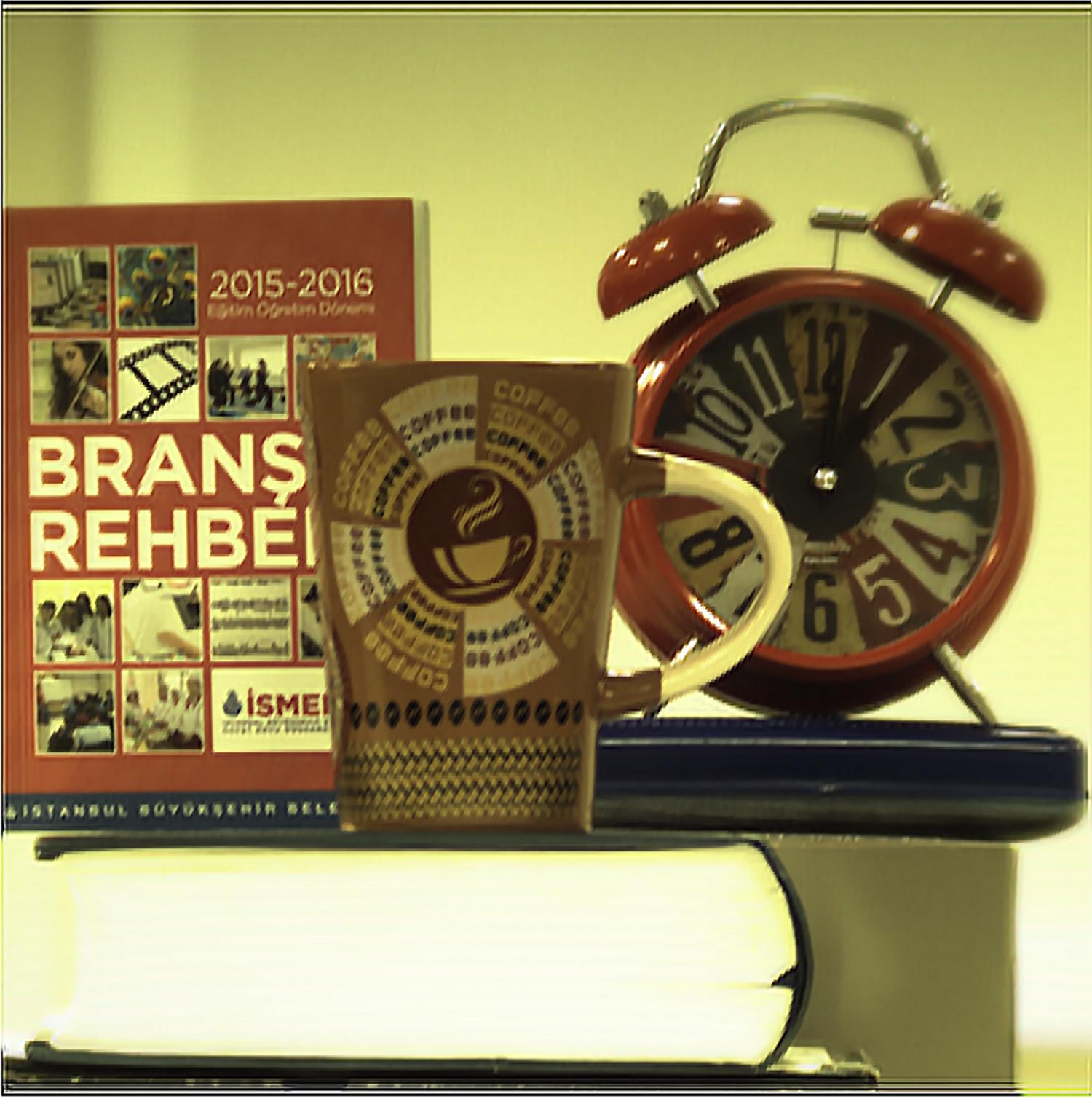}
  \includegraphics[width=2cm,trim={100 850 1100 350},clip=true,width=3.2cm,keepaspectratio]{RF_Mid_Full_deblurred.jpg}
  \includegraphics[width=2cm,trim={500 800 800 500},clip=true,width=3.2cm,keepaspectratio]{RF_Mid_Full_deblurred.jpg}  
  \\
  \rotatebox{90}{~~~~~~~~Super-resolution}
  \includegraphics[width=4cm]{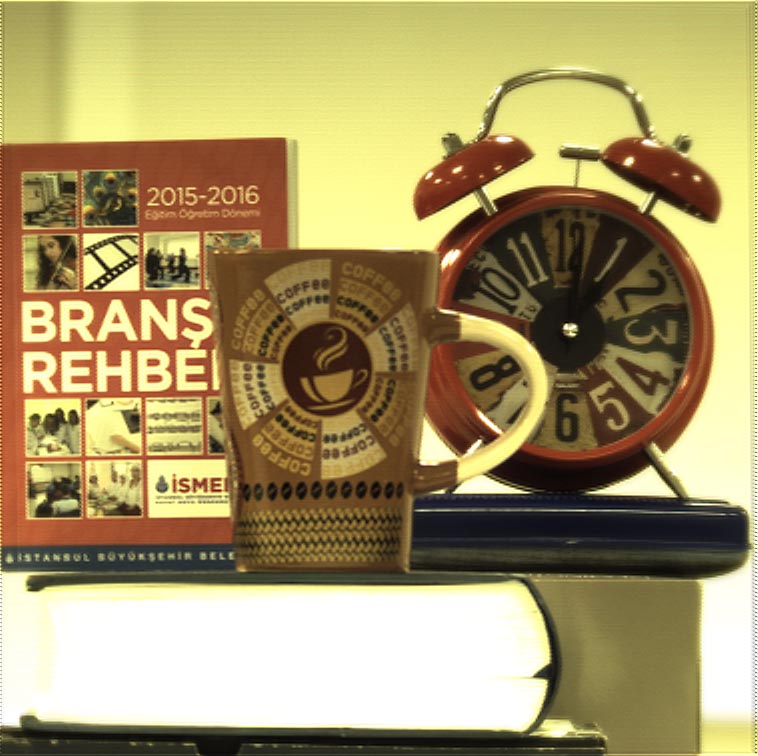}
  \includegraphics[width=2cm,trim={50 425 550 175},clip=true,width=3.2cm,keepaspectratio]{SteepestDescent1_0.jpg}
  \includegraphics[width=2cm,trim={250 400 400 250},clip=true,width=3.2cm,keepaspectratio]{SteepestDescent1_0.jpg}  
  \\
  \rotatebox{90}{~~~~~~~~Proposed}
  \includegraphics[width=4cm]{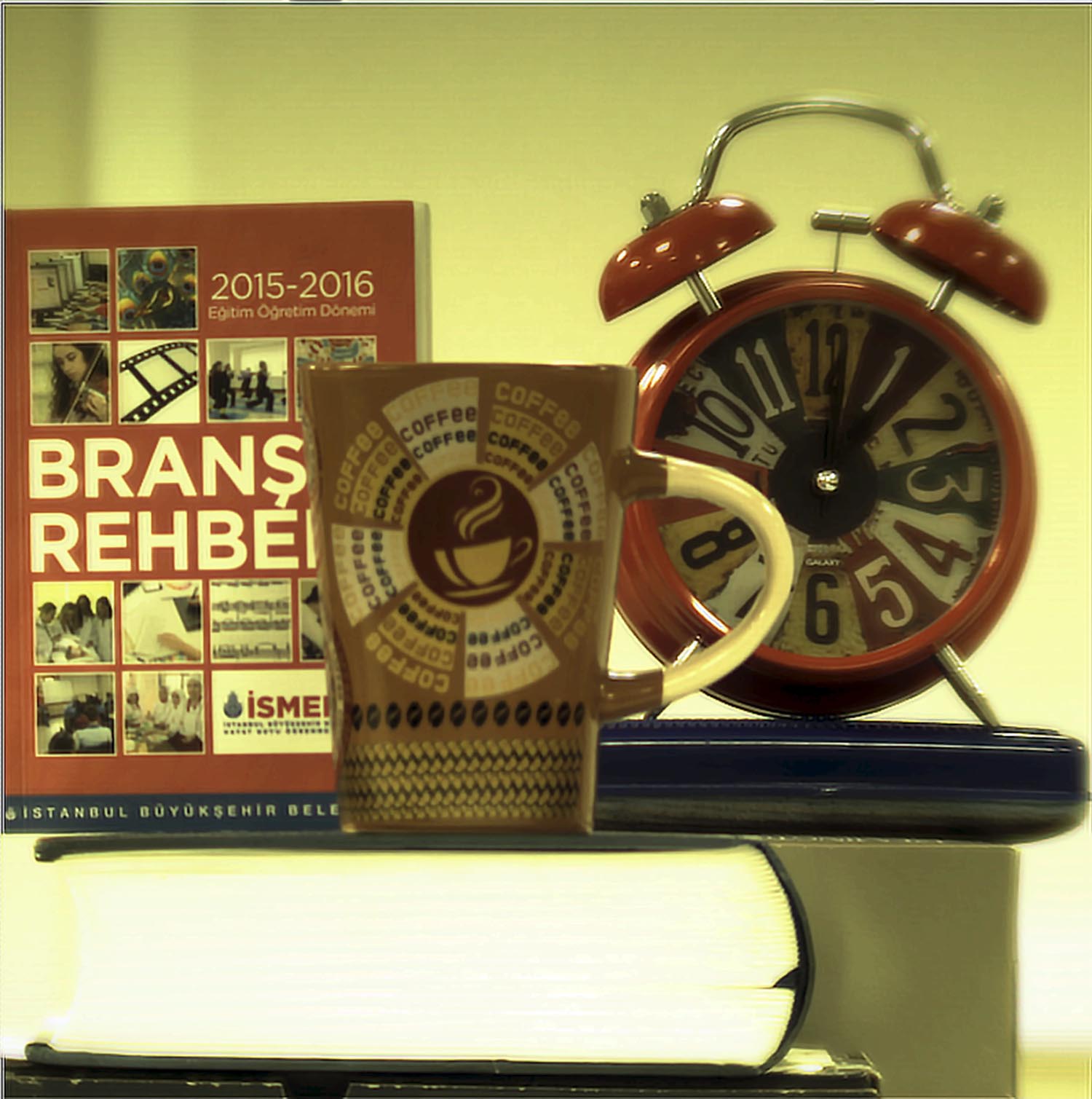}
  \includegraphics[width=2cm,trim={100 850 1100 350},clip=true,width=3.2cm,keepaspectratio]{RF_MidH_Full_deblurred.jpg}
  \includegraphics[width=2cm,trim={500 800 800 500},clip=true,width=3.2cm,keepaspectratio]{RF_MidH_Full_deblurred.jpg}
  \caption{Comparison of the proposed method (with deconvolution) against bicubic interpolation (before and after deconvolution) and super-resolution restoration of a single light field.}
\label{fig:superresolution}
\vspace{-1em}
\end{figure*}

In Figure \ref{fig:superresolution}, we compare the proposed method against super-resolution restoration. The super-resolution method that we use is a standard one, specifically, Bayesian estimation with a Gaussian prior \cite{Gunturk2006}. The input images to the super-resolution process are the perspective images of a light field; that is, there are $7\times7=49$ input images. In the figure, it is seen that while the super-resolution restoration improves over bicubic interpolation, it cannot achieve the resolution enhancement that the proposed method does through micro-scanning.

\begin{figure}[h]
\centering
  \includegraphics[width=5.2cm]{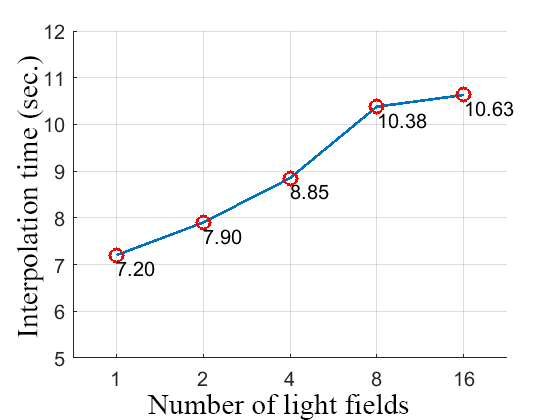}
  \caption{Time required to generate single high resolution light field perspective versus the number of input light fields. The deconvolution process takes an additional time of about 7 seconds.}
\label{fig:Dataset1:InterpTime}
\vspace{-1em}
\end{figure}

Finally, we investigate the computational cost of the proposed method. The implementation is done with MATLAB, running on a PC with Intel Core i5-4570 processor clocked at 3.20 GHz with 12 GB RAM. The computation times for different number of input light fields (per perspective) are shown in Figure \ref{fig:Dataset1:InterpTime}. For the deconvolution process, we use executable software provided by the authors of the paper \cite{xu2010two}; the software takes about 7 seconds to deconvolve a perspective image.  

\section{Conclusions}
\label{sec:Conclusion}

In this paper, we presented a micro-scanning based light field spatial resolution enhancement method. The shift amounts are determined for a specific light field sensor according to the arrangement and size of micro-lenses. The method incorporates a shift estimation step to improve the accuracy of the micro-shifts. With a translation stage of high accuracy, the registration step could be eliminated. 

The method assumes that the scene is static, relative to the amount of time it takes to capture the light fields. This is indeed the main assumption of all micro-scanning based resolution enhancement applications. One possible extension of micro-scanning techniques to applications with dynamic scenes is to incorporate an optical flow estimation step into the process. We did not investigate this idea as it is outside the scope of this paper. 

We should also note that software-based light field super-resolution methods are complimentary to the micro-scanning based idea presented here. That is, it is possible to apply a software based super-resolution method to the light field obtained by micro-scanning to further increase the light field spatial resolution.



\end{document}